%% file: main.tex
\documentclass[10pt,twocolumn,letterpaper]{article}

\usepackage{iccv}
\usepackage{times}
\usepackage{epsfig}
\usepackage{url}            % simple URL typesetting
\usepackage{booktabs}       % professional-quality tables
\usepackage{amsfonts}       % blackboard math symbols
\usepackage{nicefrac}       % compact symbols for 1/2, etc.
\usepackage{microtype}      % microtypography
\usepackage{amsthm,amsmath,amssymb} % define this before the line numbering.
\usepackage{color}
\usepackage{subfigure}
\usepackage{multicol}
\usepackage{multirow}
\usepackage{graphicx}
\usepackage{algorithm} %format of the algorithm 
\usepackage{algpseudocode}
\usepackage{mathrsfs}
\usepackage[pagebackref=true,breaklinks=true,letterpaper=true,colorlinks,bookmarks=false]{hyperref}

\iccvfinalcopy % *** Uncomment this line for the final submission

\ificcvfinal\fi

\begin{document}

%%%%%%%%% TITLE
\title{Adaptive Pixel-wise Structured Sparse Network for Efficient CNNs}

\author{Chen Tang\\
Tsinghua University\\
{\tt\small tangc20@mails.tsinghua.edu.cn}
\and
Wenyu Sun\\
Tsinghua University\\
%%First line of institution2 address\\
{\tt\small wy-sun16@mails.tsinghua.edu.cn}

\and
Zhuqing Yuan\\
Tsinghua University\\
%%First line of institution2 address\\
{\tt\small yuanzhuqing@tsinghua.edu.cn}

\and
Yongpan Liu\\
Tsinghua University\\
%%First line of institution2 address\\
{\tt\small ypliu@tsinghua.edu.cn}
}
\maketitle
\ificcvfinal\thispagestyle{empty}\fi

%%%%%%%%% ABSTRACT
\begin{abstract}
To accelerate deep CNN models, this paper proposes a novel spatially adaptive framework that can dynamically generate pixel-wise sparsity according to the input image.
The sparse scheme is pixel-wise refined, regional adaptive under a unified importance map, which makes it friendly to hardware implementation.
A sparse controlling method is further presented to enable online adjustment for applications with different precision/latency requirements. 
% Therefore, our method supports rapid deployment on devices with various computing power.
The sparse model is applicable to a wide range of vision tasks.
Experimental results show that this method efficiently improve the computing efficiency for both image classification using ResNet-18 and super resolution using SRResNet.
On image classification task, our method can save 30\%-70\% MACs with a slightly drop in top-1 and top-5 accuracy.
On super resolution task, our method can reduce more than 90\% MACs while only causing around 0.1 dB and 0.01 decreasing in PSNR and SSIM. Hardware validation is also included.
\end{abstract}

%%%%%%%%% BODY TEXT
\input{1-introduction.tex}
\input{2-related_work.tex}
\input{3-Feature_Sparse_Regularization.tex}

\input{4-experiments.tex}
\input{5-conclusion.tex}

{\small
	\bibliographystyle{ieee_fullname}
	\bibliography{egbib}
}

\end{document}

%% file: 1-introduction.tex
\section{Introduction}
\label{sec:Intro}
\begin{figure}[htbp]
	\begin{center}
		%\fbox{\rule{0pt}{2in} \rule{.9\linewidth}{0pt}}
		\includegraphics[width=0.475\textwidth]{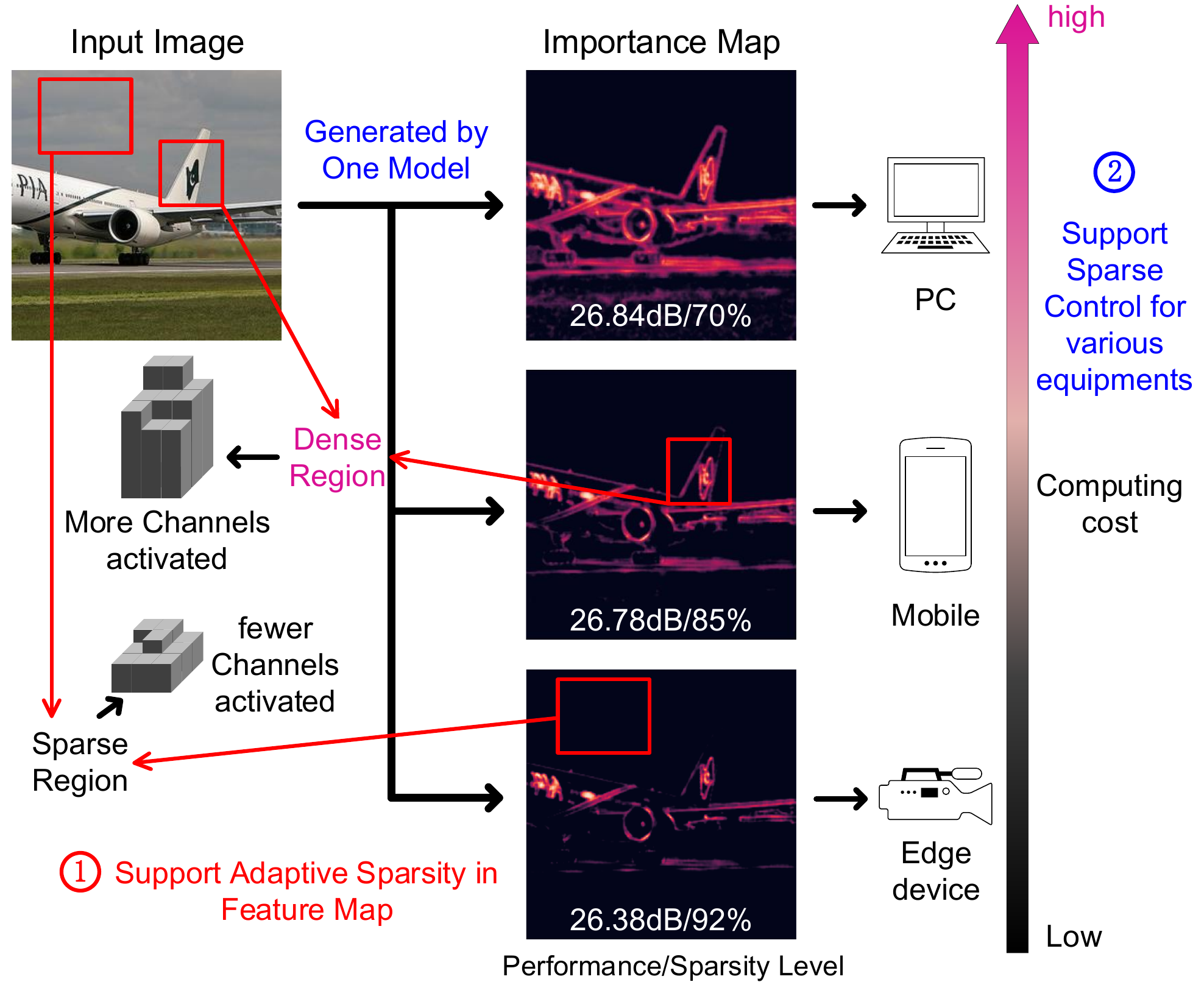}
	\end{center}
% 	\vspace{-8pt}
	\caption{Illustration of our work on image restoration task. An importance map $\mathbf{F}$ is generated from the input image to define sparsity for each pixel. Larger magnitude in $\mathbf{F}$, which displays a brighter color, means more numbers of channels are activated in the feature map for the corresponding region. The total network sparsity is controlled through importance maps by a dynamic adjustment mechanism in one model without retraining.}
	\vspace{-8pt}
	\label{fig_introduction}
\end{figure}

Convolution neural networks (CNNs) have achieved great success for a wide range of computer vision tasks in recent years. However, the high accuracy is often achieved at the cost of heavy memory and computation consumption with deeper and wider CNNs, which prevents it to be applied to  embedded devices and real-time applications.  
%For example, the time cost to process one frame ranges from 200ms to 1500ms\cite{inproceedings} in the video super resolution tasks. 
%It is far more than the popular adopted video frame rate. 

To solve the problem, %one way is to train light-weighted models directly, which  usually causes a decreasing in accuracy. 
one way is to use weight pruning methods, which can effectively compress the weight parameters to slim the models. Early methods generate unstructured sparsity \cite{lee2018snip,han2016deep,HASSIBI1993Second,guo2016dynamic,han2015learning}, which greatly compress the model but is not convenient for hardware deployment. Therefore, more researches focus on learning structured sparsity \cite{li2016pruning,he2017channel,liu2017learning}. Although the weight pruning methods can effectively reduce the model parameters, they are static without considering the spatial redundancy of the image. They have the same computing cost for different inputs or image regions, and cannot make adaptation spatially, which limits the further improvement of model accuracy.  

Recently, dynamic model inference has attracted more attention. They attempt to dynamically allocate the computing resources according to specific requirements and limitations. These methods are input-dependent, which means the computing cost is adaptively changed for different images, even though the image size keeps the same. They can properly allocate more computing resources for the importance regions such as the ROI (Region of Interest) pixels, while pay less attention to the sub-important regions such as the background. Researches \cite{figurnov2017spatially, li2017not, Veit2018Convolutional} first propose several layer-wise methods to activate different layers for different inputs or distribute different numbers of layers for important and sub-important regions. Then several channel-wise \cite{lin2017runtime,gao2018dynamic} methods are proposed, which can dynamically empty the one whole channel to save computation.
% Weight pruning methods do not see such coarse-grained design while these kinds of design can work for dynamic model inference, which to some extent shows the potential of dynamic design philosophy. Effectiveness of these methods are verified on tasks like image classification and semantic segmentation. 
% While methods of this kind are most easy for hardware implementation, the performance needs extra improvement. 
%These methods are usually adopted in the segmentation tasks where the size of feature map is unchanged when passing convolution layers. When the size of feature map is becoming smaller with pooling layer as in classification task, the layer-wise method is not compatible because the operation of early stop should change the feature map size before the final fully-connected layer.
To further improve the performance, the more refined pixel-wise \cite{dong2017more, ren2018sbnet, xie2020spatially, kuen2018stochastic, kong2017ron} methods are proposed. The philosophy is to save only important pixels to compose the feature map. However, the selected pixels tend to be different between layers, which exerts extra overhead for online and offline compiling and not very friendly for hardware implementation. Our method is also pixel-wise based while performs in a different manner. Firstly, we treat pixels differently according to their importance level instead of just discarding those unimportant ones. Pixels in sub-important region can be activated with fewer channels while the important ones more. In this way, our method can maintain better accuracy/latency trade-off. Secondly, The sparsity is determined by an unified importance map, which makes the structure of sparsity invariant from layer to layer. This makes our method easier to deploy on hardware devices than other pixel-wise methods.
Sometimes, it could be necessary to deploy models at different sparse levels on the same device. For example, when the battery is well charged, the model can run with the full use of computing resources. When the battery is running out, models with higher sparse level are needed. Also, different applications can have different power or latency requirements. Previous methods need to train many models offline to cope with this problem, which is inconvenient and time-consuming. Through special design, our method can automatically control the sparse level online in one model, which can be a better solution to cope with the heavy burden of multiple models training.
    The main contributions of this paper include:
% Experiments show that up to 60\%-95\% sparsity can be generated for super resolution tasks and 15\%-50\% channels can be generated for image classification tasks.
% We show that: 1) With the application of our method, the results can be comparable with those of the non-sparse one. Our method can cut the multiply-accumulate operations dramatically. Up to 95\% Macs can be saved for super resolution tasks with fewer than 0.2dB PSNR drop, and a 1.3-2X mac saving can be achieved for image classification task with less than 1\% top5 accuracy drop. 2)We put forward a method to train one model that supports sparsity control. Therefore, our method can support rapid deployment for different embedded devices. The main contributions of this paper include: 
\begin{itemize}
\item [1)]  We propose a spatially adaptive CNN model to dynamically generate pixel-level sparsity and reduce the storage and computing cost. The sparsity is guided by a unified importance map. An example of hardware implementation is also presented.
\item [2)]  Our method can online control the sparsity level in one model, thus saving the burden to train multiple models for various devices or requirements. 
\item [3)]  Our methods can be widely utilized on various vision tasks. We validate the effectiveness on the super resolution and image classification tasks. Experimental results show the proposed models achieve comparable or better performance with state-of-art methods.
\end{itemize}

%% file: 2-related_work.tex
\section{Related Work}
\label{sec:related_work}
Our method is related with various previous works like structured sparsity pruning and input-dependent execution. In this section, we briefly introduce static weight pruning and dynamic execution. In weight pruning, network sparsity is generated by pruning the unimportant weights, which results in a static sparse network. While in dynamic execution, the sparsity is generated in an input-dependent manner.

\subsection{Static Weight Pruning} Weight pruning is very effective in reducing redundant connections in the model \cite{lee2018snip,han2016deep,HASSIBI1993Second,guo2016dynamic,han2015learning}. In particular, structured pruning \cite{li2016pruning,he2017channel,liu2017learning} has gained more and more attention, since the pruned network is more friendly to hardware deployment. Structured pruning can be further divided into filter pruning, channel pruning and kernel shape pruning according to the specific methods \cite{wen2016learning}.

\textbf{Filter pruning} searches for unimportant filters, by pruning which the corresponding output channels can be skipped to generate sparsity \cite{he2019filter,li2016pruning,luo2017thinet,ding2019centripetal}. 
The filters can be selected based on the $l_1$-norm of weight channels \cite{li2016pruning}, the statistics information computed from its following layer \cite{luo2017thinet} or the correlation between each filters \cite{ding2019centripetal}. Filter pruning can largely reduce the model size but may cause accuracy loss because of its large-grained nature.

\textbf{Channel pruning} searches for unimportant channels to leave out. Thus, filters producing this channel and those related ones in the following layer can be pruned  \cite{liu2017learning,he2017channel,lin2019towards,zhuang2018discrimination,ye2018rethinking}. The pruned channel can be selected based on LASSO regularization \cite{liu2017learning}, reconstruction error before and after pruning \cite{he2017channel}, generative adversarial learning \cite{lin2019towards} or an attention mechanism \cite{zhuang2018discrimination}.

\textbf{Kernel Shape pruning} or pattern pruning reduces convolution kernels patterns to a limited number, which also generates structured sparsity in kernel-wise. Ma \textit{et al.} \cite{ma2020pconv} propose pattern pruning and force all the 3X3 convolution kernels collapsing into several patterns which only has four non-zero parameters. Wang \textit{et al.} \cite{wang2020high} select patterns based on probability density function and design a hardware architecture to better support the algorithm. 

Although weight pruning can reduce the model size, its sparsity is static in network inference regardless of the input data. Therefore, some researchers turn to the dynamic execution methods, of which the generated sparsity is input-dependent. 

%Filter pruning and channel pruning methods view tensor as a box of regular shape. If the channel or filter is rated as unimportant, all the pixels will lose the information provided by this channel or filter. For kernel shape pruning, all the pixels are processed by the same filters consist of the limited kernel patterns. Thus the weights are Static in these sparse models regardless of the input data. 

\subsection{Dynamic Sparsity Aware Execution}
Dynamic Execution methods try to utilize the redundancy existed in the feature map to generated sparsity dynamically. Dynamic execution is further divided into layer-wise, channel-wise and pixel-wise methods.
%Many works is orthogonal to methods above and often cannot actually compress the model.

\textbf{Layer-wise} dynamic execution methods explore to reduce the number of layers to relieve computation burden according to input data \cite{figurnov2017spatially, li2017not, Veit2018Convolutional}. 
%Pruning weight layers are thought to be too coarse-grained in static model pruning. However, dynamically skipping layer is considered a successful trial. 
The skipping decision is made based on a halting score that is generated by the input of current layer \cite{figurnov2017spatially}, the classifying difficulty estimated from pixels \cite{li2017not} or a gate function attached with each layer \cite{Veit2018Convolutional}. 

\textbf{Channel-wise} dynamic execution methods dynamically inactivate channels so that calculations related with them can be saved. Lin \textit{et al.} \cite{lin2017runtime} use Markov decision process to dynamically prune the channel based on the input. Gao \textit{et al.} \cite{gao2018dynamic} uses an attention mechanism to dynamically choose the top-k importance channels to preserve in each layer. 

Layer-wise and channel-wise methods skip the computation related to specific layers or channels, which take the sparsity of the input as a whole or average While it is easy to notice that different inputs have different spatial sparsity. Thus pixel-wise methods are proposed to fill this gap. 
%Our method is different from them in that we do not discard layers. All the pixels in our method will go through same number of layers with different number of channels active. Therefore, we think our method are more fine-grained.

\textbf{Pixel-wise} methods try to let fewer pixels involved in the calculation based on the spatial redundancy \cite{dong2017more, ren2018sbnet, xie2020spatially, kuen2018stochastic, kong2017ron}. One approach is to sample representative or important pixels and process them only. The sampling is done by separating images into background and foreground blocks \cite{ren2018sbnet} or by a gumbel-softmax based stochastic sampling module \cite{xie2020spatially}. Another idea is to reduce the number of pixels by down-sampling feature map \cite{kuen2018stochastic}. 
% Although there often exists pixel-wise sparsity in the input data, omitting pixels directly is believed to be coarse-grained. 
The omitted pixel is prone to be scattered in the former methods. Changing the size of the feature map also limits the application in some tasks, such an super resolution, or segmentation. In this work, we propose a pixel-wise structured sparsity network that different numbers of channels are activated for different pixels. Also, our method support online sparse control, which previous methods cannot offer.

%% file: 3-Feature_Sparse_Regularization.tex
\begin{figure*}[h]
	\begin{center}
		%\fbox{\rule{0pt}{2in} \rule{.9\linewidth}{0pt}}
		\includegraphics[width=1\textwidth]{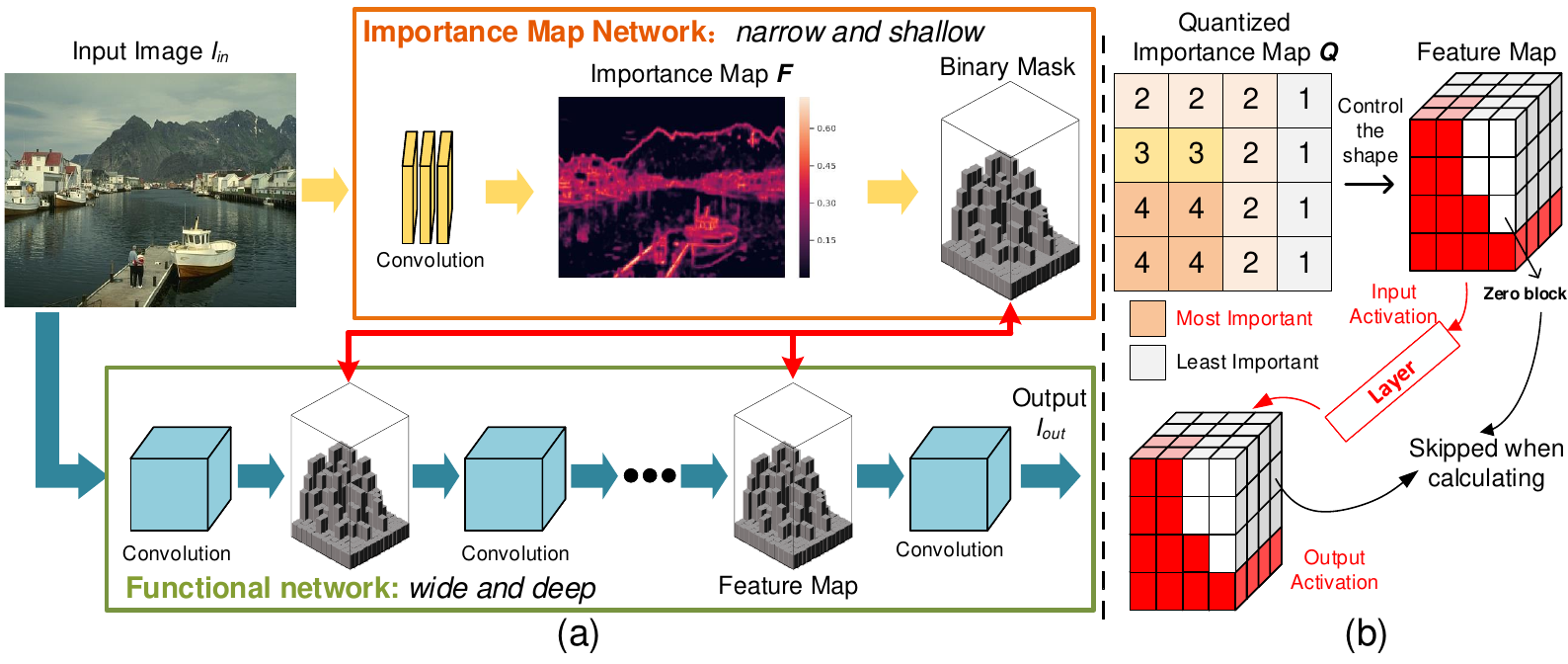}
	\end{center}
	%\vspace{-7pt}
	%\centering
% 	\vspace{-5pt}
	\caption{(a)Overview of Our Method. The mask is first generated and all the intermediate feature maps will share the same shape with it (b) A Detailed View of Activation Sparsity. The quantization mechanism is exlained in Section.~\ref{section_imn}. Activation Sparsity is the sparsity along the channel dimension.}
% 	\vspace{-5pt}
	\label{fig:overview}
\end{figure*}

\section{Proposed Method}
\label{sec:proposed method}
The overall framework of the proposed spatially adaptive network is shown in Fig.~\ref{fig:overview}(a), and is composed of two parts: an importance map network and a main functional network. The former one generates an importance map $\mathbf{F}$, which indicates the importance level of each pixel in the input image. The network is designed to be extremely light-weight. $\mathbf{F}$ is quantized to represent the number of valid output channels for each pixel. The main functional network represents various CNN solutions in a wide range of vision tasks, such as image classification or super-resolution network. Fig.~\ref{fig:overview}(b) illustrates the sparse inference progress. The sparsity in feature map is fully controlled by a unified quantized importance map $\mathbf{Q}$ from $\mathbf{F}$, and zero blocks can only be in place of latter channels. As a result, the proposed spatially adaptive sparse network is structured for hardware implementation.

\subsection{Importance Map Network}
\label{section_imn}
The importance map network (IMN) aims at generating a sigmoid activated feature map $\mathbf{F}^{ H \times W \times 1}$ to indicate the important regions of the image, which can be illustrated in Eq.~\ref{eq:IMN}, where $\psi()$ refers to the IMN module.
\begin{equation}
\label{eq:IMN}
\mathbf{F} = \psi(I_{in}), \quad \mathbf{F}(i,j) \in (0,1)
\end{equation}
% With the input image $I_{in} \in \mathbb{R}^{H \times W \times 3}$ getting through the layers and the sigmoid activated function, an
% , an input image $I_{in} \in \mathbb{R}^{H \times W \times 3}$ pass $\textbf{n}$ convolution layers to get the sigmoid activated importance map $\mathbf{F}^{ H \times W \times 1}\in (0,1)$ can be generated. 
Regions with lower value in $\mathbf{F}$ are less important, where fewer channels will be allocated to these pixels.
% Pixels with higher value in $\mathbf{F}$ indicates that the pixel is in a more important region and more channels need to be allocated. 
To realize such spatial sparsity in the feature map, a binary mask $\mathbf{M}$ is generated from the quantized $\mathbf{F}$ in the training phase.

Specifically, we first quantize the $\mathbf{F}(i,j)$  into $\mathbf{Q}(i,j)$ by $\mathrm{L}$ levels with quantization step $\mathrm{S}= \lfloor \mathrm{C}/\mathrm{L} \rfloor$, where $\mathrm{C}$ is the number of channels in the original dense feature map. 
\begin{equation}
\label{eq:1}
\mathbf{Q}(i,j) = l-1, \quad \frac{l-1}{\mathrm{L}} \leq \mathbf{F}(i,j) < \frac{l}{\mathrm{L}}, \quad l=1,...,\mathrm{L}
\end{equation}
Then we generate the binary mask $\mathbf{M}$ according to the quantized $\mathbf{Q}$. For each pixel $\mathrm{m} = \mathbf{M}(i,j)\in \mathbb{R}^{1\times C}$, it can be denoted as Eq.~\ref{eq:2}. With higher $\mathbf{Q}(i,j)$, more channels for pixel $(i,j)$ are activated. 
\begin{equation}
\label{eq:2}
\mathrm{m}(k) = \left\{
\begin{aligned}
& 1, k \leq  \mathrm{S} \times \mathbf{Q}(i,j)   \\
& 0, k >     \mathrm{S} \times \mathbf{Q}(i,j) \\
\end{aligned}
\right.
\end{equation}

Since the operation of quantization in Eq.~\ref{eq:1} is non-differential, we adopt a \textbf{tanh} function to approximate the gradient propagation, which enables the whole network can be end-to-end trained. As a result, Eq.~\ref{eq:1} and Eq.~\ref{eq:2} can be combined together as Eq.~\ref{eq:3}. The \textbf{sign} function binarizes the input into \{0,1\} and keeps the gradients from output to input unchanged. $\alpha$ is a scale factor to control the gradient magnitude, which is set for 4 in our paper. During training, $\mathbf{M}$ makes dot product with the output of each convolution layer to force the feature map to be sparse. The zero blocks are skipped to reduce the computing costs in inference phase.
\begin{equation}
\label{eq:3}
\small
\mathrm{m}(S \times l: S\times l+S) = \textbf{sign}(\textbf{tanh}(\alpha\times( \mathbf{F}(i,j)-\frac{l}{L})))
\end{equation} 

\subsection{Integrated with different CNN blocks}
\begin{figure}[htbp]
\begin{center}
   \includegraphics[width=0.95\linewidth]{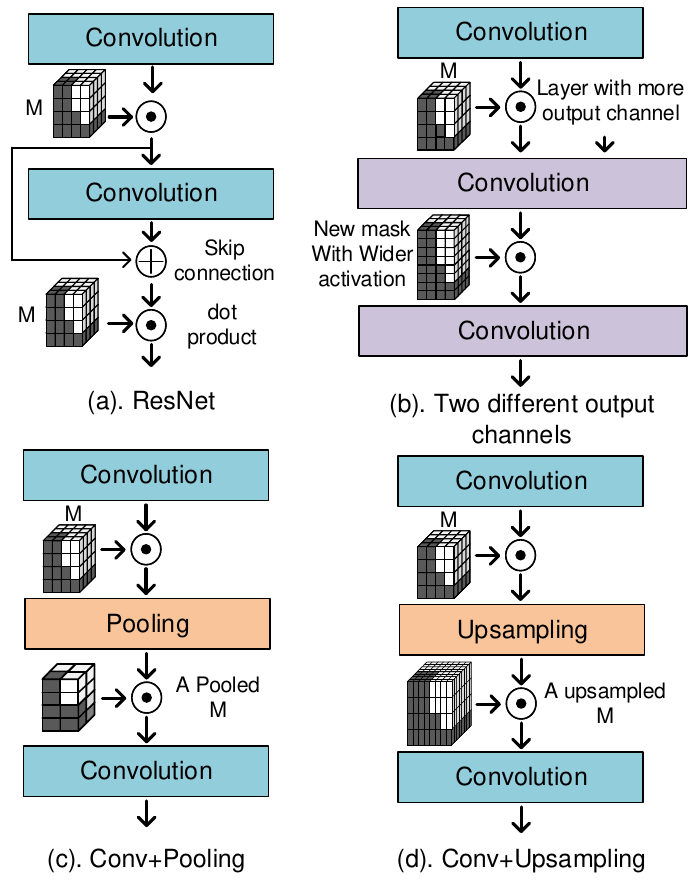}
\end{center}
   \caption{Integrating binary mask $\mathbf{M}$ with different CNN blocks.}
\label{fig:mask}
\vspace{-8pt}
\end{figure}

The computed binary mask $\mathbf{M}$ can be integrated into multiple kinds of CNN modules to generate sparsity in the feature map. Fig.~\ref{fig:mask} shows how to generate $\mathbf{M}$ in multiple common cases during training. In ResNet, all convolution outputs make dot product with $\mathbf{M}$ after the skipping connection. In This way, the sparse structure in the output feature map will not be destroyed. In layers that input and output have different numbers of channels, two different masks $\mathbf{M_1}$ and $\mathbf{M_2}$ are needed. They can both be quantized from the importance map $\mathbf{F}$ ahead. 
% with the same level of $\mathrm{L}$ but different step $\mathrm{S}$ because of different channel number $\mathrm{C}$. 
In some cases, the spatial scales of the feature map may change, like the up-sample operation in super resolution tasks or pooling layers in classification tasks. To ensure $\mathbf{M}$ can have identical shape,  
% In an up-sample operation that is usually adopted in super-resolution task or a convolution-pooling block that is usually  adopted in classification task, since the size of feature map can be changed, the size of importance map $\mathbf{F}$ and the binary mask $\mathbf{M}$ should also be adjustable. 
one choice is to resize $\mathbf{F}$ through extra convolution layers. Another solution is to directly up-sampling (e.g., nearest interpolation) or pooling (e.g., max pooling) $\mathbf{F}$ to make the spatial size compatible with the feature map. In real implementation, we find the latter method can achieve better performance with less computing cost. The claim will be discussed in the section of ablation study.      

\subsection{Spatially Sparse Controlling}
\label{subsec:sparse-controlling}
Based on the trainable IMN module, the next question is how to successfully train a valid importance map $\mathbf{F}$. The training of $\mathbf{F}$ relies on our loss function, which will be first presented.
% In addition to optimizing the original loss of the main functional network, we first add a regularization term by means of $\mathbf{F}$ to enforce the feature map to be sparse.
Then we further consider an optimizing method to make the model sparsity controllable, where implementation of our Automatic Sparse Controlling (ASC) module and the change of the loss function will be introduced.

\textbf{Loss function.} The loss function $\mathcal{L}_{\textrm{total}}$ is denoted in Eq.~\ref{eq:4}, which is composed by two parts: the original loss function $\mathcal{L}_{\textrm{main}}$ for the main functional network and the regularization term $ \mathcal{L}_{\textrm{sparse}} = \vert l_1(\mathbf{F})-\gamma\vert$ from the IMN module, where $l_1$ is the $l_1$-norm function. $\mathcal{L}_{\textrm{main}}$ can be the cross entropy loss for classification task, the L1/L2 loss for the super-resolution task, or some others. The $\mathcal{L}_{\textrm{sparse}}$ aims to make
% makes constraint on the importance map $\mathbf{F}$ or the quantized map $\mathbf{Q}$ in Eq.~\ref{eq:2} to make 
the model sparse. Optimizing $\mathcal{L}_{\textrm{main}}$ and $\mathcal{L}_{\textrm{sparse}}$ is an adversarial process. The balanced point is controlled by the hyper-parameter $\lambda$ and $\gamma$.
% , of which the latter one represents the targeted sparse level. 
The final model tends to be more sparse with larger $\lambda$ and smaller $\gamma$. With such defined $\mathcal{L}_{\textrm{total}}$, the two network can be optimized together to trade off the accuracy and the sparse level properly and adaptively. In this way, we can train a model without ASC module. 
\begin{equation}
\label{eq:4}
\mathcal{L}_{\textrm{total}}=\mathcal{L}_{\textrm{main}} + \lambda \mathcal{L}_{\textrm{sparse}}
\end{equation} 

\textbf{Automatic Sparse Controlling.} Pruning multiple models with various sparse levels is inconvenient and time-consuming when the accuracy/latency requirements change. 
% Noticing that the sparsity is in fact directly decided by the quantized importance map $\mathbf{Q}$ from Eq.~\ref{eq:2},  
A straightforward way is to manually increase or decrease each value in $\mathbf{Q}$ to adjust the the number of activated channels. However, it is hard to decide which pixel should be changed to promise the final accuracy. Therefore, an ASC module is designed to learn where to adjust $\mathbf{Q}$ is better and how to control the amount of change in sparse level. 

\begin{figure}
	%\vspace{-7pt}
	\begin{center}
		%\fbox{\rule{0pt}{2in} \rule{.9\linewidth}{0pt}}
		\includegraphics[width=0.475\textwidth]{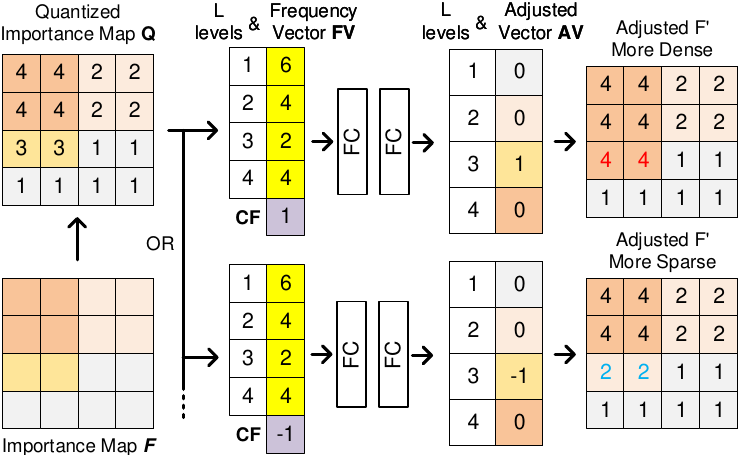}
	\end{center}
% 	\vspace{-6pt}
	\caption{Illustration of the automatic sparse controlling method.}
	\vspace{-4pt}
	\label{fig:multi-sparsity}
\end{figure}

Since the importance level of these pixels has already been measured by the IMN module, there is no need to reinvent the wheel by adjusting $\mathbf{Q}$ pixel by pixel. In addition, $\mathbf{Q}$ is quantized from $\mathbf{F}$ and it can be observed that pixels with same quantization levels are prone to gather together. Therefore, we try to do same change to pixels with same value in $\mathbf{Q}$ (e.g. add them by one for all). In order to control the amount of total change in sparsity, the frequency of occurrence for each quantization level is needed. Also, an extra signal called control factor ($\mathbf{CF}$) that indicates the change direction is considered.
The total procedure is shown in Fig.~\ref{fig:multi-sparsity}. Firstly, the statistical frequency for each value is counted to compose a 1-D tensor denoted as the frequency vector $\mathbf{FV}$. Then, an adjusted vector $\mathbf{AV}$ is generated from the concatenation of $\mathbf{FV}$ and $\mathbf{CF}$ to indicate the adjusting weight for different $\mathrm{L}$ regions. The total adjustment in sparse level and the process can be formulated by Eq.~\ref{eq:FV2AV}.
\begin{equation}
\label{eq:FV2AV}
\begin{aligned}
\Delta l_1(\mathbf{F}) &= l_1(\mathbf{F'}) - l_1(\mathbf{F})= \mathbf{FV^{T}}\cdot \mathbf{AV}\\
&= \mathbf{FV^{T}}\cdot\varphi(\mathbf{FV},\mathbf{CF})
\end{aligned}
\end{equation} 
% The final adjustment is determined according to the $\mathbf{AV}$. 
During training, we randomly generate an integer $\mathbf{CF}$, e.g., $\mathbf{CF}\in \{-2,-1,0,1,2\}$. Each $\mathbf{CF}$ represents each sort of sparse adjustment. To make the adjustment controlled by $\mathbf{CF}$, the $\mathcal{L}_{\textrm{sparse}}$ in this mode can be redefined in Eq.~\ref{eq:5}, where $\beta$ means the targeted sparse change. $\beta$ is controlled by $\mathbf{CF}$, e.g., $\beta=0.1$ for $\mathbf{CF}=1$ and $\beta=-0.1$ for $\mathbf{CF}=-1$. In this way, we can directly choose different values of $\mathbf{CF}$ in inference phase to online obtain different sparse levels. The total procedure is summerized in Algorithm~\ref{al1}. We show in ablation study that the ASC module can effectively adjust the sparse level online and perform better the straightforward method. 
\begin{equation}
\label{eq:5}
\mathcal{L}_{\textrm{sparse}} = \vert l_1(\mathbf{F})-\gamma \vert + \vert\Delta l_1(\mathbf{F}) - \beta\vert
\end{equation}

\begin{algorithm}
	\caption{Sparse Controlling Flow}
	\label{al1}
	\begin{algorithmic}[1]
% 		\Require
% 		 $W_{\psi}$ and $W_{\varphi}$: parameters from IMN module, ASC , : parameters from functional network.
% 		\Ensure
% 		$W_{f}$: updated parameters from Importance Map Network, $W_{\varphi}$: updated parameters from functional network. \\
% 		initialize $W_{f}\leftarrow \widehat{W}_{f}$, $W_{\varphi}\leftarrow \widehat{W}_{\varphi}$ and iter $\leftarrow$ 0.
        \State $\mathbb{S}$: set of $\mathbf{CF}$. $\psi()$: IMN network. $\varphi()$: ASC network. $\phi()$: main functional network. $\mathbf{FP}$: feature map
		\Repeat 
		\State Fetch a training mini-batch x
		\State $\mathbf{F}=\psi(x)$, $\quad$ $\mathbf{Q}$ = Quantize($\mathbf{F}$)
		\State Randomly choose $\mathbf{CF}$ from $\mathbb{S}$, $\quad$ $\beta \leftarrow \mathbf{CF}$
		\State $\mathbf{FV} \leftarrow \mathbf{Q}$,$\quad$ $\mathbf{AV}$ = $\varphi(\mathbf{FV})$
		\State Generate adjusted $\mathbf{F'}$ from $\mathbf{F}$ and $\mathbf{AV}$
		\State $\mathbf{M} \leftarrow \mathbf{F'}$
		\State $\mathcal{L}_{\textrm{main}}$ = loss($\phi(\mathbf{FP \cdot M})$, labels)
		\State Generate $\mathcal{L}_{\textrm{sparse}}$ by Eq.~\ref{eq:FV2AV} and \ref{eq:5}
		\State $\mathcal{L}_{\textrm{total}}=\mathcal{L}_{\textrm{main}} + \lambda \mathcal{L}_{\textrm{sparse}}$
		\State Back propagation and update parameters
		\Until 
		\State Iters reaches desired maximum
	\end{algorithmic}
\end{algorithm}

%% file: 4-experiments.tex
\section{Experiment}
\label{sec:experiment}
Our methods can be utilized on multiple applications. Super resolution (SR) and Image classification tasks are chosen to validate our design. The size of the feature maps in SR network is gradually increased while it is reduced for the classification network. Therefore, the two tasks are representative enough to include all common-used CNN blocks.

\begin{figure}[h]
% 	\vspace{-6pt}
	\begin{center}
		%\fbox{\rule{0pt}{2in} \rule{.9\linewidth}{0pt}}
		\includegraphics[width=0.45\textwidth]{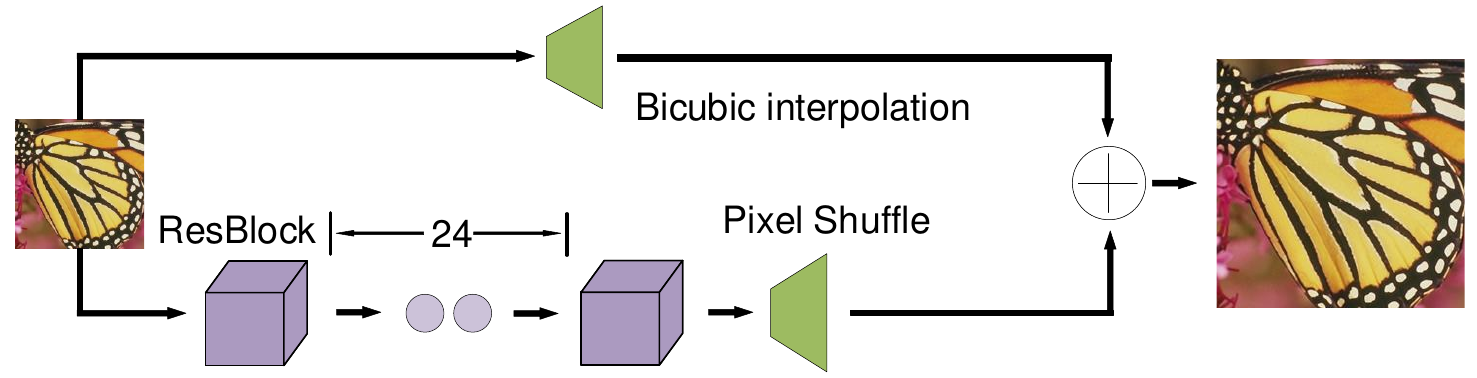}
	\end{center}
% 	\vspace{-8pt}
	\caption{Baseline Model for Super Resolution}
	\vspace{-8pt}
	\label{Res-SR}
\end{figure}
\begin{table*}[hbtp]
\label{tab:SRresult}
	\caption{Comparsion between our models with other state-of-the-art light-weight methods for SRx2. The best and sub-best performance is highlighted in red and blue. MACs is computed in average when the original HR images are 720P}
	\begin{center}
	\resizebox{0.8\textwidth}{!}{
	\begin{tabular}{c|c|c|c|c|c|c|c|c|c}
		\hline
		\hline
		\multirow{2}*{Method}  & \multirow{2}*{MACs} & \multicolumn{2}{|c|}{Set5} & \multicolumn{2}{|c|}{Set14} & \multicolumn{2}{|c|}{B100} & \multicolumn{2}{|c}{Urban100}  \\ \cline{3-10}
		&&PSNR & SSIM & PSNR & SSIM & PSNR & SSIM & PSNR & SSIM \\
		\hline
		FSRCNN\cite{dong2016accelerating}         & \color{red}6.0G       & 37.00     &0.9558        & 32.63       &0.9088        &31.53        &0.8920         &29.88       &0.9020 \\
		DRRN\cite{tai2017image}           & 6796.9G    & 37.74     &0.9591        & 33.23       &0.9136        &32.05        &0.8973         &31.23       &0.9188 \\
		MemNet\cite{tai2017memnet}         & 623.9G     & 37.78     &0.9597        & 33.28       &0.9143        &32.08        &0.8978         &31.31       &0.9195 \\
		SelNet\cite{choi2017deep}         & 225.7G     & 37.89     &0.9598        & 33.61       &0.9160        &32.08        &0.8984         &-           &-      \\
		CARN\cite{ahn2018fast}           & 222.8G     & 37.76     &0.9590        & 33.52       &0.9166        &32.09        &0.8978         &31.92       &0.9256 \\
		OISR-RK2-s\cite{he2019ode}     & 316.2G     & 37.98     &0.9604&\color{blue}33.58    &0.9172        &32.18        &0.8996&\color{blue}32.09    &0.9281 \\
		OISR-LF-s      & 316.2G&\color{red}38.02&\color{blue}0.9605&\color{red}33.62     &0.9178        &32.20        &\color{blue}0.9000&\color{red}32.21     &0.9290 \\
%		Mask-SR1*   & X2   & 354G  &\color{blue}38.01&\color{blue}0.9606&33.44   &0.9178&\color{red}32.22&\color{red}0.9005   &31.61&\color{blue}0.9342 \\
%		%Mask-SR2*   & X2   & 349G       & 38.00     &\color{red}0.9607&33.46     &0.9177&\color{blue}32.21&\color{blue}0.9004 &31.60&\color{red}0.9345 \\
		Ours($\lambda$=1e-4)      & 352G       & 37.95     &\color{blue}0.9605        & 33.41       &\color{blue}0.9180\color{blue}&\color{blue}32.21    &\color{blue}0.9000         &31.42       &\color{blue}0.9325 \\
		Ours($\lambda$=5e-4)      & 209G       & \color{blue}38.00&\color{red}0.9606 & 33.38       &\color{red}0.9182\color{red}&\color{red}32.22     &\color{red}0.9002         &31.49       &\color{red}0.9328 \\
		Ours($\lambda$=2.5e-3)      & \color{blue}30.5G      & 37.88     &0.9599        & 33.30       &0.9164        &32.11        &0.8985         &31.21       &0.9284 \\
		\hline
% 		Q1(CF=-1) & 51.2G &37.88&0.9597&33.33&0.9160&32.11&0.8981&31.32&0.9285\\
% 		Q1(CF=0)  & 78.1G &37.95&0.9601&33.40&0.9169&32.16&0.8991&31.44&0.9309\\
% 		Q1(CF=1)  & 138.8G &37.97&0.9604&33.42&0.9174&32.18&0.8996&31.48&0.9319\\
% 		Q1(CF=2)  & 216.9G &37.99&0.9605&33.42&0.9176&32.18&0.8998&31.49&0.9323\\ \hline
		Ours-ASC(CF=0)  & 72.2G &37.90&0.9599&33.36&0.9165&32.14&0.8984&31.36&0.9293\\
		Ours-ASC(CF=1)  & 112.5G &37.95&0.9602&33.42&0.9171&32.17&0.8993&31.47&0.9317\\
		Ours-ASC(CF=2)  & 174.5G &37.97&0.9604&33.44&0.9175&32.19&0.8996&31.51&0.9324\\
		Ours-ASC(CF=3)  & 262.1G &37.99&0.9605&33.45&0.9178&32.20&0.8998&31.52&0.9328\\
		\hline
		MSRN\cite{li2018multi}          & 1356.8G    & 38.08     &0.9605        & 33.74       &0.9170        &32.23        &0.9013         &32.22       &0.9326 \\
		Baseline        & 1770.6G    & 38.12     &0.9611        & 33.60       &0.9194        &32.27        &0.9010         &31.92       &0.9374 \\   
		\hline
		\hline       
	\end{tabular}}
	\end{center}
 	\vspace{-6pt}
\end{table*}
\begin{table}[htbp]
	\label{tab:class}
	\caption{Comparsion between our model and other pruning or execution methods in ResNet-18.
	}
% 	\vspace{-12pt}
	\begin{center}
		\resizebox{0.475\textwidth}{!} {
		\begin{tabular}{c|c|c|c|c}
			\hline
			\hline
			Method & Type & Top1 Acc & Top5 Acc & Mac Saving\\ \hline
			SFP\cite{he2018soft} & weight pruning & 3.18$\downarrow$& 1.85$\downarrow$ & 1.72X\\
			NS\cite{liu2017learning}  & weight pruning & 1.77$\downarrow$ & 1.29$\downarrow$ & 1.39X\\
			DACP\cite{zhuang2018discrimination} & weight pruning & 2.29$\downarrow$ & 1.38$\downarrow$ & 1.89X\\
			FPGM\cite{he2019filter} & weight pruning & 1.87$\downarrow$ & 1.15$\downarrow$ & 1.72X\\
			LCCL\cite{dong2017more} & pixel-wise & 3.65$\downarrow$ & 2.30$\downarrow$ & 1.53X\\
			CGNN\cite{hua2018channel} & layer-wise & 1.62$\downarrow$ & 1.03$\downarrow$ & 1.61X\\
			FBS\cite{gao2018dynamic} & channel-wise & 2.54$\downarrow$ & 1.46$\downarrow$ & 1.98X\\			
			ours($\lambda=1e-1$) & pixel-level & 0.14 $\uparrow$ & 0.44$\uparrow$ & 1.39X\\
			ours($\lambda=2.5e-1$) & pixel-level  & 0.75 $\downarrow$& 0.2 $\downarrow$& 1.85X\\
			ours($\lambda=3e-1$) & pixel-level  & 1.41$\downarrow$ & 0.74$\downarrow$ & 2.08X\\
			\hline
			ours-ASC(CF=1) & pixel-level & 0.90$\downarrow$ & 0.25$\downarrow$ & 1.62\\
			ours-ASC(CF=0) & pixel-level  & 0.95$\downarrow$ & 0.32$\downarrow$ & 1.75\\
			ours-ASC(CF=-1) & pixel-level  & 1.12$\downarrow$ & 0.37$\downarrow$ & 1.84\\
			\hline
			\hline
		\end{tabular} }
	\end{center}
	\vspace{-20pt}
\end{table}
\begin{table}[htbp]
	\caption{Different structures of importance map network}
% 	\vspace{-6pt}
	\begin{center}
		\label{tab:fmap_ab}
		\resizebox{0.47\textwidth}{!}{
			\begin{tabular}{c|c|c|c|c}
				\hline
				\hline
				\multirow{2}*{Structure of IMN module} & \multirow{2}*{Mac Saving} &\multicolumn{3}{|c}{results for PSNR}\\\cline{3-5}
				&& Set5 & Set14 & BSD100 \\\hline
				6 Conv, 32 channel (8.7G) & 22.0X & 37.91 & 33.42 & 32.17\\
				4 Conv, 32 channel (4.5G) & 22.6X & 37.91 & 33.41 & 32.16\\
				4 Conv, 64 channel (17.5G)& 18.5X & 37.92 & 33.47 & 32.20\\
				3 Conv, 64 channel (9.0G) & 16.7X & 37.99 & 33.49 & 32.20\\
				\hline
				\hline
		\end{tabular}}
	\end{center}
 	\vspace{-18pt}
\end{table}
\subsection{Super Resolution}
\textbf{Experimental setup.} We apply our method on the classic single image super resolution (SISR) task and the baseline model is SRResNet \cite{ledig2017photo}. We set adequate 24 Res-Blocks and increase the channel number to 128 in intermediate layers, which results in more than 1000G operations of multiply and accumulate (MACs) for the dense model. The spatially sparse models using proposed method are trained on the widely used DIV2K \cite{timofte2017ntire} dataset for the 2X scaling up task. The results are measured on four standard benchmarks, including Set5 \cite{bevilacqua2012low}, Set14 \cite{zeyde2010single}, BSD100 \cite{martin2001database} and Urban100 \cite{huang2015single}.  Peak signal to noise ratio (PSNR) and structural similarity index (SSIM) \cite{wang2004image} are adopted as the evaluation criteria. The binary mask $\mathbf{M}$ is applied to every intermediate feature map except for the last layer that generates the 3-channel output images. Adam optimizer and 2e-4 learning rate is adopted for all models. We set $\lambda=2.5e-3$ and let the sparse controlling factor $\mathbf{CF}$ to be randomized in $\{0, 1, 2, 3\}$ during training to support automatic sparse controlling within one model. The $\gamma$ is set for 0.2 in Eq.~\ref{eq:5}. Several models without the equipment of ASC module are also trained, which use different $\lambda$ settings. We compare our method with several light-weight SR networks \cite{dong2016accelerating, tai2017image, tai2017memnet,choi2017deep,he2019ode,li2018multi}

\textbf{Test results} 
%We first directly integrated with the importance map network without the ASC (Automatic Sparse Controlling) module to train 3 sparse models with different $\lambda \in \{1e-4,5e-4,2.5e-3\}$ in loss function defined by Eq.~\ref{eq:4}. Fig.~\ref{fig:sr_3model} shows the experimental results for such 3 sparse models in the PSNR-MACs scatter plot.
The final reconstruction accuracy and computing costs are listed in Table.~\ref{tab:SRresult} in various datasets. When compared to the baseline model, the proposed method can dramatically save the computation cost by reducing more than 90\% MACs while only causing around 0.1 dB decreasing in PSNR and less than 0.01 SSIM drop. The model tested with four controlling factors $\mathbf{CF}$ achieves comparable or even better performance with fewer computing costs. We are better for SSIM on most of the conditions. A clear PSNR/MACs trade-off is shown on Fig.~\ref{fig:ASC}. Our model also supports ultra efficient inference when  $\mathbf{CF}$ reach its least while the PSNR or SSIM are still promised. Therefore, it can be adopted for real-time applications to reduce latency. Moreover, our method can be rapidly deployed on various embedded systems with different computing power without retaining new models. The $\mathbf{CF}$ can control the sparse level easily, which is convenient for freely development.

% \begin{figure}[htbp]
% 	%\vspace{-7pt}
% 	\begin{center}
% 		%\fbox{\rule{0pt}{2in} \rule{.9\linewidth}{0pt}}
% 		\includegraphics[width=0.475\textwidth]{visio_figures/exp/SR.pdf}
% 	\end{center}
% % 	\vspace{-8pt}
% 	\caption{Result of directly integrated with the importance map network for SRResNetx2 tested on BSD100. }
% % 	\vspace{-7pt}
% 	\label{fig:sr_3model}
% \end{figure}
% It can also be seen that the range of sparsity level can be easily controlled by the setting of $\mathbf{CF}$, 

% \begin{figure}[htbp]
% % 	\vspace{-3pt}
% 	\begin{center}
% 		\includegraphics[width=0.45\textwidth]{visio_figures/exp/mul-new.pdf}
% 	\end{center}
% % 	\vspace{-8pt}
% 	\caption{Results for Multi-sparsity Control from One Model tested on BSD100}
% % 	\vspace{-3pt}
% 	\label{fig:ASC}
% \end{figure}

\subsection{Image Classification}
\textbf{Experimental setup.} We further validate our method on another typical vision task of image classification. ResNet-18 \cite{he2016deep} is set as the baseline model. We train the models on the official ImageNet dataset \cite{russakovsky2015imagenet}. The baseline ResNet-18 are retrained to have top1 and top5 error as 30.23\% and 10.91\%. We train 200 epochs for each model. Since there exists 2 stride convolution layers in ResNet-18, we adopt max pooling operation on the adjusted importance map $\mathbf{F}_{\mathrm{adjusted}}$ for different sizes of feature maps. For ResNet-18, we generate 4 different important maps $\mathbf{F}_{\downarrow 2}$, $\mathbf{F}_{\downarrow 4}$, $\mathbf{F}_{\downarrow 8}$, and $\mathbf{F}_{\downarrow 16}$. The final $L_{sparse}$ is defined by a average sum of $l_1$-norm on these four maps. 
% We first directly integrated with the importance map network without the ASC module to train 3 sparse models with different $\lambda \in \{1e-1,2.5e-1,4e-1\}$ in loss function defined by Eq.~\ref{eq:4}. 
We set $\lambda=2.5e-1$ and the sparse controlling factor $\mathbf{CF}$ can be randomized from $\{-1,0,1\}$ during training to support 3 levels of sparsity. Several models without ASC module is also trained with different $\lambda$ settings. We compare our methods with other pruning or dynamic execution methods. \cite{he2018soft,liu2017learning,zhuang2018discrimination,he2019filter,dong2017more,hua2018channel,gao2018dynamic}. 
\begin{figure*}[htbp]
% 	\vspace{-3pt}
	\begin{center}
		\includegraphics[width=1\textwidth]{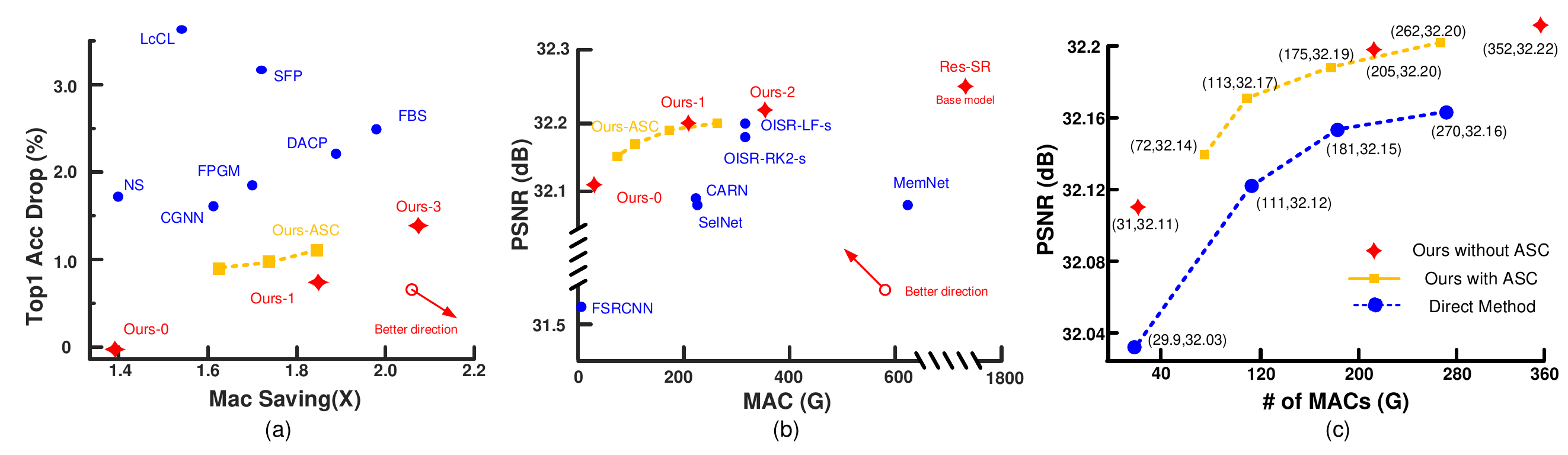}
	\end{center}
 	\vspace{-6pt}
	\caption{(a) TOP1 Acc Drop/MAC trade-off for classification tasks (b) PSNR/MAC trade-off tested on BSD100 (c) Ablation study for ASC module on super resolution tested on BSD100}
% 	\vspace{-6pt}
	\label{fig:ASC}
\end{figure*}

\textbf{Test results.} The results can be seen in Table~\ref{tab:class}. By applying our method, only a slightly top-1 and top-5 accuracy drop with a 1.4-2.1X MACs saving can achieved. Our method can achieve better results than previous fixed weight pruning methods or dynamic execution methods. A clear Top1 accuracy drop/MAC Saving trade-off is shown on Fig.~\ref{fig:ASC}. The proposed model has refined pixel-wise sparsity to improve the accuracy compared with the layer-wise sparse model CGNN \cite{hua2018channel} and pixel-wise model LCCL \cite{dong2017more}. Also, our method can effectively provide the online sparse control service.

% As a result, we set the summing weight  on loss larger for the former feature maps with larger size. A fair average way is  also conducted as a comparison in Table.~\ref{tab:weight}. It can been seen that both method can train a model with more sparsity in the former feature maps with larger size. The weighted method can obtain a more  sparse model with higher accuracy compared with the average method, which valid that the former feature maps can be more sparse in pixel level. 
% \begin{table}[htbp]
% 	\caption{Comparison between model that $l_{\mathrm{sparse}}$ is a weighted sum or  a fair average  of $\mathbf{F}$ in different sizes.
% 	}
% % 	\vspace{-8pt}
% 	\begin{center}
% 		\label{tab:weight}
% 		\resizebox{0.475\textwidth}{!}{
% 			\begin{tabular}{c|c|c|c|c|c|c}
% 				\hline
% 				\hline
% 				\multirow{2}*{Model} & \multicolumn{4}{|c|}{Mac Saving} &\multicolumn{2}{|c}{Acc Drop}\\\cline{2-7}
% 				&$\downarrow 2$ &$\downarrow 4$ &$\downarrow 8$ &$\downarrow 16$ &Top1 &Top5\\\hline
% 				Weighted & 7.53X & 2.55X & 1.43X & 1.16X & 2.51 & 1.50\\\hline
% 				Avg & 4.22X & 2.63X & 1.78X & 1.38X & 2.94 & 1.94\\
% 				\hline
% 				\hline
% 		\end{tabular}}
% 	\end{center}
% % 	\vspace{-6pt}
% \end{table}
\subsection{Ablation Study}
\subsubsection{Importance Map Network}
We do ablation study on the structure of importance map network. For our previous experiments, the super resolution task has four convolution layers with 32 channels in between and the classification task has three convolution layers where it first expands to 18 layers and then comes up to 36 layers. We claim that the design of IMN module can be light-weight so that it will only cause little extra latency. We train multiple models with different structures. All the contrast experiments are done on the same hyper parameters settings. As it can be seen in Table.~\ref{tab:fmap_ab}, different structures of IMN module can lead to little difference for final results. Therefore, there is no need to search the best structure. The latency caused by IMN module is negligible compared to the main functional network.
% Therefore, the design of importance network will not be a bottleneck for the whole performance.
\subsubsection{Automatic Sparse Controlling Network}
Normally, models trained with specific sparse level can have better performance than the one that can online generate multiple sparse levels. It would be unworthy if the accuracy/latency trade-off deteriorate much with the integration of ASC module. The contrast results can be seen on Fig.~\ref{fig:ASC}. Our ASC module can maintain the performance with the bonus of automatic sparse controlling, which validate its worthiness. The final PSNR seems no difference for super resolution. Accuracy on classification may have some little drop while the performance is still promising. In addition, to validate the effectiveness of our design, we compare our ASC module to another sparse control method, which directly increase or decrease numbers of channels for all the feature map. Fig.~\ref{fig:ASC} shows that the proposed ASC method can be more effective to improve the final model accuracy by more than 0.05 dB compared with the native way. A possible explanation for the success is that our ASC method takes the statistical distribution of pixel importance into account and the learning way helps it to automatically find the best adjusted distribution to trade off between the sparsity and accuracy properly. 
% \begin{table}[htbp]
% 	\caption{Ablation study for different down-sampling strategies.}
% % 	\vspace{-6pt}
% 	\begin{center}
% 		\label{tab:downsample}
% 		\resizebox{0.47\textwidth}{!}{
% 			\begin{tabular}{c|c|c|c|c|c|c|c}
% 				\hline
% 				\hline
% 				\multirow{2}*{Strategies} & \multicolumn{5}{|c|}{Mac Saving for different sizes} &\multicolumn{2}{|c}{Acc Drop}\\\cline{2-8}
% 				&$\downarrow 2$ &$\downarrow 4$ &$\downarrow 8$ &$\downarrow 16$ &Total &Top1 &Top5\\\hline
% 				MaxPooling & 1.48X & 1.22X & 1.12X & 1.06X& 1.20X & 0.59 & 0.43\\
% 				%Max2 & 2.22X & 1.40X & 1.16X & %1.07X & 1.05 & 0.65\\
% 				AvgPooling & 1.23X & 1.23X & 1.23X & 1.23X &1.23X& 0.97 & 0.69\\
% 				%Avg2 & 1.41X & 1.23X & 1.23X & 1.14X & 0.90 & 0.69\\
% 				\hline
% 				\hline
% 		\end{tabular}}
% 	\end{center}
% 	\vspace{-16pt}
% \end{table}
\begin{figure}[htbp]
	\begin{center}
		%\fbox{\rule{0pt}{2in} \rule{.9\linewidth}{0pt}}
		\includegraphics[width=0.47\textwidth]{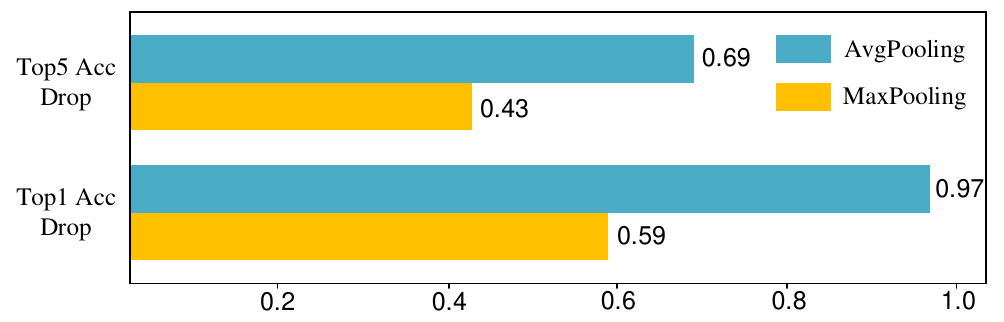}
	\end{center}
	%\vspace{-7pt}
	%\centering
	\vspace{-6pt}
	\caption{Ablation study for different down-sampling strategies.}
 	\vspace{-6pt}
	\label{fig:downsample}
\end{figure}
\begin{figure*}[h]
	\begin{center}
		%\fbox{\rule{0pt}{2in} \rule{.9\linewidth}{0pt}}
		\includegraphics[width=0.95\textwidth]{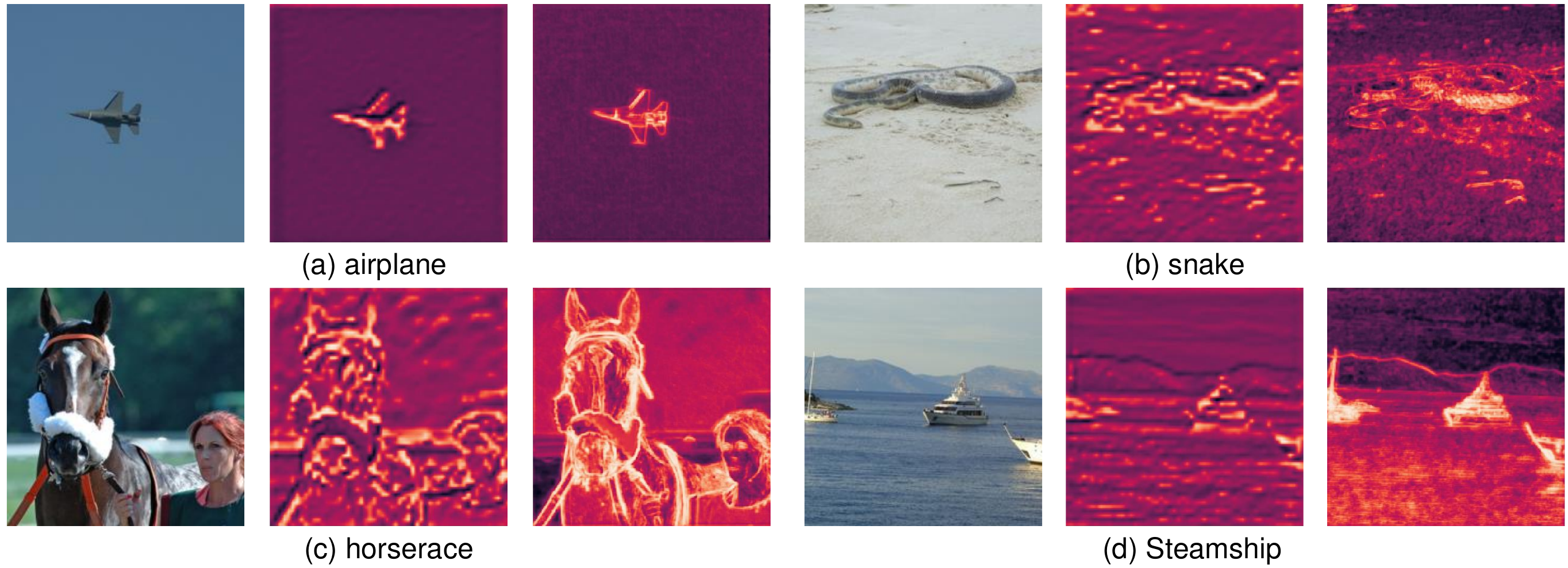}
	\end{center}
	%\vspace{-7pt}
	%\centering
% 	\vspace{-7pt}
	\caption{Input and their corresponding importance map from different tasks under similar sparsity level. The left one is from the image classification task and the right one comes from super resolution.}
 	\vspace{-5pt}
	\label{fig:imp_map}
\end{figure*}
\subsubsection{Integration Strategy}
When the feature map is reshaped, our $\mathbf{M}$ should have a corresponding adaptation. As is mentioned in the methodology section, most of the conditions can be handled easily except the one $\mathbf{F}$ needs to be down-sampled. Since the feature map encounters four down-sampling in ResNet-18, we do our ablation study on the image classification tasks. we study three down-sampling methods: max pooling, average pooling, and stride-2 convolutions. 
We find that the former two methods can be much easier trained than the latter one. One possible explanation is that the stride-2 convolution will generate several importance maps with different distributions, thus it is difficulty to find the best weights on these maps as they are summed up together to form the $\mathcal{L}_{\textrm{sparse}}$. Therefore, the latency/accuracy trade-off tends to be worse than the former two methods.
% the $l_1$-norms of these importance maps are need to be summed up together as the final sparse loss. 
% In this way, we find it failed to optimize various distributions of $\mathbf{F}$ together, which results in a low accuracy performance. 
Fig.~\ref{fig:downsample} shows that max pooling method can achieve better performance than average pooling ones at similar sparse level. Average pooling makes the sparse level identical for all feature maps, while the max pooling makes smaller feature map more dense, which can keep more information in the latter layers. It can be inferred that the latter smaller feature maps are more important than the former ones to affect the accuracy. Also, since there exists little spatial redundancy when the size of feature map becomes small (e.g., 14$\times$14 and even 7$\times$7), applying pixel-wise sparsity could meet more difficulty. Therefore, max pooling is adopted in this paper.
% This may because our sparsity is in pixel level of the feature map, and the latter down-scaled map could only have a few pixels (e.g., 14$\times$14 and even 7$\times$7). So the spatially sparse method may not be suitable for such small feature maps. However, as the channel numbers have increase (e.g, 512 and even 2048), weight pruning in channel level should takes more effect. 
\begin{table}[htbp]
	\caption{Hardware validation on Intel Core i5-9600K and ZYNQ-7035 with number 0 as benchmark for ResNet-18}
% 	\vspace{-6pt}
	\begin{center}
		\label{tab:realspeedup}
		\resizebox{0.47\textwidth}{!}{
			\begin{tabular}{c|c|c|c|c|c|c}
				\hline
				\hline
				 \multirow{2}*{}& \multirow{2}*{MACs} & Theo & \multicolumn{2}{c|}{Runtime(ms)} &\multicolumn{2}{c}{Real Speedup}\\ \cline{4-7}
				 &&Speedup&i5-9600K&Zynq-7035&i5-9600K&Zynq-7035\\\hline
				0 &  1.8G  & -     & 1450 & 85.2 & - & -\\
				1 &  0.74G & 2.45X & 735  & 39.0 & 1.97X & 2.18X\\
				2 &  0.60G & 3.01X & 575  & 31.1 & 2.52X & 2.73X\\
				3 &  0.48G & 3.73X & 461  & 24.6 & 3.15X & 3.46X\\
				% \multicolumn{5}{c}{Benchmark tested for ResNet-18 with 1.45s time cost}\\
				\hline
				\hline
		\end{tabular}}
	\end{center}
	\vspace{-16pt}
\end{table}
% \begin{table}[htbp]
% 	\caption{Realistic speedup on ZYNQ-7035. Benchmark tested for ResNet-18 with 85ms}
% % 	\vspace{-6pt}
% 	\begin{center}
% 		\label{tab:realspeedup-fpga}
% 		\resizebox{0.47\textwidth}{!}{
% 			\begin{tabular}{c|c|c|c|c}
% 				\hline
% 				\hline
% 				 & MACs & Theoretical speedup & run time & realistic speedup\\ \hline
% 				1 &  0.78G & 2.45X & 39.0ms & 2.18X\\
% 				2 &  0.63G & 3.01X & 31.1ms & 2.73X\\
% 				3 &  0.50G & 3.73X & 24.6ms & 3.46X\\
% 				% \multicolumn{5}{c}{Benchmark tested for ResNet-18 with 1.45s time cost}\\
% 				\hline
% 				\hline
% 		\end{tabular}}
% 	\end{center}
% 	\vspace{-16pt}
% \end{table}
\subsection{Hardware Validation}
% Good theoretical latency/accuracy trade-offs are achieved through our method. However, the realistic speedup is more important.
% To validate the effectiveness of the proposed method, we test the sparse network in both CPU and FPGA. For CPU implementation, we rebuild the whole resnet-18 in C code and test the inference time on Intel Core i5-9600K. For FPGA implementation, we adopt the HLS technology by Xilinx to design a hardware accelerator especially supporting such kind of sparsity.  We consider the sparse coding and scheduling technique to minimize the off-chip memory access and maximize the utilization of PE array.  Table. 4 shows the theoretical and realistic performance improvement. It can be seen that the realistic speedup is closed to the theory for both CPU and FPGA. There is only a little decreasing caused by the overhead of importance map network. 
We test the speedup of three sparse level on Intel Core i5-9600K and the results are shown in Table.~\ref{tab:realspeedup}. For fair comparison, we replace all the layers by our C code based implementations and used four threads for one running model. Since the index of zero blocks is known in advance, we can easily skip them to save computation. 
% Same philosophy has been taken by LCCL \cite{dong2017less}, as they predict where should be zero in the latter layers and neglect them. 
Our hardware implementations is simple, as the structure of our sparsity ensures that we only need the first index of zero blocks. Although the I/O bottleneck and the complexity to utilize the sparsity of the input of one layer make the realistic speedup lower than the theoretical ones, our method remain good performance. We also design a FPGA-based accelerator using ZYNQ-7035 device and high-level synthesis (HLS) technology by Xilinx. ResNet-18 is also chosen to be the benchmark and the realistic speedup is close to the theoretical ones. The detailed description about the deployment method on FPGA is beyond the scope of this paper. Further information may be found in \cite{sun2020112}.
% Although the sparse algorithm has been proved to be effectiveness, it is unrealizable to accelerate the model inference without specific hardware support for such kind of sparsity in feature map. So in addition to the algorithm research, we also develop a hardware architecture that fully support the spatially adaptive sparse model in FPGA. Owing to the structured sparsity decided by the importance map $\mathbf{F}$ only, a sparse scheduler method is proposed to skip the zero blocks for each input and output pixel and a importance map guided sparse decoder is adopted to minimize the memory access.  The FPGA-based accelerator can achieve highest 765 GOPS/W energy efficiency in the extreme sparse level. The detailed information is beyond the scope of this paper, but we provide a demonstration video for $\times 2$ SR in supplementary materials.  
\subsection{Algorithm Discussion}
We generate the importance map from various tasks, as is shown in Fig.~\ref{fig:imp_map}. It can be seen that they both highlight the important region of the input image. The difference is caused by the different goals of the tasks. Super resolution aims to reconstruct detailed textures. Therefore, the importance map also shows which region is hard to transform into the high resolution ones. More channels are activated in those areas to help reconstruction. Importance map
in the image classification tasks works in a different way. It samples the most representative image patches to extract more semantic information. Although the importance image seems not as elaborate as the ones in super resolution tasks, we can see a refined highlight of the interested object in the image.

%% file: 5-conclusion.tex
\section{Conclusion}
\label{sec:conclusion}
In this paper, we propose a novel spatially adaptive method to generate structured sparsity in CNN-based models. We adopt a pixel-wise sparse method that is dynamically adapted by an unified importance map, which achieves good performance to reduce the network computing cost while maintaining the accuracy. An adjustable mechanism to online control the sparse level in one model is also developed to avoid training multiple models for specific applications. Our method can be widely used on various CNN structure and is validated on two different vision tasks: super resolution and classification. Experimental results show that our approach achieves high performance in both of them. Hardware validation is also implemented to prove the practical value of our method.

%\section{Broader Impact}
%\label{sec:broader}
%Our Feature Regularization can be combined with other modules easily. We just need to generate a corresponding $mask$ of the input data batch and apply multiplication with intermediate feature batches, as shown in Fig\ref{general-structure}. It can reduce the computation burden because there is no need to calculate at the position where channels are discarded and the storage burden will be also relieved. Our work is inspired from Mu Li’s Content-weighted Image Compression\cite{li2018learning}, which increases compression rate by applying different Quantization bits to different pixels. However, We have different training strategy and usage. We identify that our FR has potential for all the pixel to pixel tasks. It contributes to the re-simplifying of each data patch at the pixel level. In addition, we design a method for multi-sparsity model, which also differs our work from them.

%Traditional methods focus on pruning weights and generating sparsity, which lacks flexibility for different kinds of inputs. Our method can provide a different view from pruning channels and can generate sparsity according to the input image. However, we only test its effectiveness for super resolution problems. Whether other pixel to pixel problems can apply our FR remains in the theoretical stage. As more and more algorithms design networks that have tremendous channels, FR can mend the consequence of huge number of parameters brought by such design concept.